\documentclass{Interspeech2024}

\interspeechcameraready 

\usepackage{multirow}
\usepackage{algpseudocode}
\usepackage{algorithm}
\usepackage{comment}
\usepackage{pgfplots}
\pgfplotsset{compat=1.18}

\title{Sequential Editing for Lifelong Training of Speech Recognition Models}

\name[]{Devang}{Kulshreshtha}
\name[]{Saket}{Dingliwal}
\name[]{Brady}{Houston}
\name[]{Nikolaos}{Pappas}
\name[]{Srikanth}{Ronanki}

\address{
  AWS AI Labs, USA
\email{\{kulshrde, skdin, hstbrady, nppappa, ronanks\}@amazon.com}
}


\begin{document}

\maketitle

\begin{abstract}
    
    Automatic Speech Recognition (ASR) traditionally assumes known domains, but adding data from a new domain raises concerns about computational inefficiencies linked to retraining models on both existing and new domains. Fine-tuning solely on new domain risks Catastrophic Forgetting (CF). To address this, Lifelong Learning (LLL) algorithms have been proposed for ASR. Prior research has explored techniques such as Elastic Weight Consolidation, Knowledge Distillation, and Replay, all of which necessitate either additional parameters or access to prior domain data. We propose \textit{Sequential Model Editing} as a novel method to continually learn new domains in ASR systems. Different than previous methods, our approach does not necessitate access to prior datasets or the introduction of extra parameters. Our study demonstrates up to 15\% Word Error Rate Reduction (WERR) over fine-tuning baseline, and superior efficiency over other LLL techniques on CommonVoice English multi-accent dataset.
\end{abstract}

\section{Introduction}
Recently, the field of speech recognition (and machine learning/AI in general) has trended toward large foundational models trained on very large, diverse datasets covering many domains. Despite this trend, it is still common in industrial settings that after a foundational/base model has been initially trained, to gradually train it on new domains or categories. In multidialect ASR, for example, these two situations would be improving the performance of the model on a single dialect (or subset of dialects) and adding a previously-unseen dialect/accent to the model. Both of these goals would typically be achieved by fine-tuning the base/foundational model, possibly with the addition/substitution of some model parameters, on new training data. 

Fine-tuning often comes with a cost, which is that the model's performance on the domains seen during training can degrade due to catastrophic forgetting \cite{kirkpatrick2017overcoming}. Returning to the example of adding a new dialect/accent to a multilingual ASR model, this degradation can be troublesome if the model is expected to perform well on the new dialect \textit{and} on all previously-seen dialects. A common mitigation approach to the catastrophic forgetting observed when fine-tuning on a new dialect is to re-train the model with both the new data and the already-seen data. This, of course, can be extremely costly, especially in the era of very large models and training datasets. In addition, the previously-seen data is not always continually-available in practical settings. Lifelong learning (or continual learning) approaches have been shown to alleviate this catastrophic forgetting effect in a wide variety of machine learning models and tasks, including ASR.

The most simple lifelong learning approach is Experience Replay; when a new domain is added via fine-tuning, a subset of the original training is also included \cite{isele2018selective} (or possibly all of the training data \cite{li2022massively}). However, this comes with the obvious downside of being inefficient, as each new domain being added requires more and more replay data. Adding multi-task training objectives to encourage the model to retain information on previous domains, as in Elastic Weight Consolidation \cite{kirkpatrick2017overcoming} and Knowledge Distillation can also be effective in ASR \cite{houston2020continual} and other tasks \cite{9879287}, but they may show limited ability to scale to many new domains. Above approaches also require either additional parameters or previous domain data to mitigate Catastrophic Forgetting (CF). More recently, several studies \cite{wortsman2022model,ilharco2023editing,yadav2023tiesmerging} investigated the manipulation and/or combination of fine-tuned model parameters with base model parameters for creating multi-task models, avoiding the need to re-use data or implement more complicated multi-task training approaches. However, these methods tend to degrade in quality when applied to a large number of tasks \cite{yadav2023tiesmerging}.

\begin{figure}
    \centering
    \includegraphics[scale=0.5]{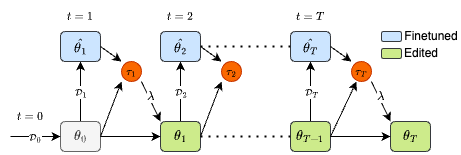}
    \caption{\textbf{Sequential Model Editing for Lifelong ASR}: At each time step $t$, the current model $\theta_{t-1}$ is fine-tuned on data $\mathcal{D}_t$ to obtain $\hat{\theta_t}$. Then task vector $\tau_t$ is computed. Finally, new model is obtained my merging $\tau_t$ with $\theta_{t-1}$ as : $\theta_t = \theta_{t-1} + \lambda \cdot (\tau_t)$.}
    \label{fig:lll-figure}
\end{figure}

In this paper, we explore model editing approaches for sequential training of ASR models that overcome limitations in prior work in terms of requiring additional parameters or access to prior domain data. Specifically, we investigate sequential editing of the original model continually trained on all the previous data sets. Here, we focus on two representative methods, namely Task Arithmetic \cite{ilharco2023editing}, which uses basic arithmetic operations to combine checkpoints from different tasks, and TIES-Merging \cite{yadav2023tiesmerging} which addresses issues that arise when task vectors (i.e. models) are combined, such as sign conflicts and small weights. These approaches have been explored for out-of-domain generalization \cite{ilharco2023editing, Jin2022DatalessKF}, multi-task learning \cite{ilharco2023editing, yadav2023tiesmerging}, and transfer learning \cite{Matena2021MergingMW} but not for lifelong learning to the best of our knowledge. At every sequential step, we assume access only on the new data source which is a challenging setting for existing continual learning methods \cite{NEURIPS2019_15825aee}. For evaluation, we focus on effectively learning multiple English dialects in an incremental fashion while preserving the performance on previously seen dialects on CommonVoice English data set. The main contributions can be summarized as follows: 

\begin{itemize}
    \item We propose a novel Sequential Model Editing approach that can be used for lifelong training of ASR models without relying on prior domain datasets or additional training and/or parameters.
    \item Our approach results in 15\% WER-reduction (WERR) over the fine-tuning baseline on CommonVoice English multi-accent dataset, compared to 6\% WERR achieved by previously proposed baselines.
\end{itemize}

\begin{algorithm}[t]
\begin{algorithmic}[1]
\footnotesize
\Require {
Data sources $\mathcal{D} = [\mathcal{D}_0, \mathcal{D}_1,\dots,\mathcal{D}_T]$, $\lambda$, and $k$.
}
\Ensure Lifelong ASR model $\theta^*$
\State \textbf{Init:} $\theta_0 \leftarrow $ Train on $\mathcal{D}_0$
\For{$t \leftarrow 1$ to $T$}
\State $\hat{\theta_t} \leftarrow $ Fine-tune $\theta_{t-1}$ on $\mathcal{D}_t$
\State $\tau_t \leftarrow \hat{\theta_t} - \theta_{t-1}$ \Comment{Task Arithmetic}
\If{TIES-Merging}
    \State $\tau_t \leftarrow$ ties\_merging\_procedure$(\tau_t, k)$
\EndIf
\State $\theta_t \leftarrow \theta_{t-1} + \lambda \cdot (\tau_t)$
\EndFor
\State $\theta^* \leftarrow \theta_T$

\end{algorithmic}
\caption{Sequential Model Editing for Lifelong ASR}\label{alg:model-editing}
\end{algorithm}

\section{Lifelong Learning for ASR}






%
ASR systems often consist of a Conformer-based CTC model \cite{gulati2020conformer, graves2006connectionist} that takes in audio sequence and outputs a text sequence. These models are often trained with paired audio and text data. 

Let $\theta_0$ be the parameters of such a model that is trained on a large set of such audio-text pairs represented by $\mathcal{D}_0$. In many practical scenarios, the abilities of the model are expanded by targeting new domains/accents/languages. More formally, let $\mathcal{D}_1, \ldots, \mathcal{D}_T$ be a sequence of $T$ data sources that are incrementally used to update the model with each dataset targeting a particular domain. As the capabilities of the ASR model expand to these new domains sequentially, it is desirable to retain the performance on the older domains. Also, with the recent use of massive volumes of datasets for training, it often becomes prohibitively difficult to store and maintain all the data sources. Similarly, for some publicly available models, training datasets are not released or are behind pay-walls, thereby making it challenging to adapt the model to new domains without catastrophic forgetting. 

To address these practical challenges, we define the goal for Lifelong Learning (LLL) for ASR as learning the optimal model parameters $\theta^*$ that performs well on all the data sources $\mathcal{D}_0 \ldots \mathcal{D}_T$, where the data sources are obtained sequentially and at any time step $t$, only the data source $\mathcal{D}_t$ is accessible and no past or future data sources. This constraint makes the existing trivial multi-task solutions unusable, which assume access to all domains data simultaneously to train the model. Therefore, we propose a novel sequential model-editing based approach as summarized in the Algorithm \ref{alg:model-editing} and Figure \ref{fig:lll-figure}.



\section{Sequential Model Editing}\label{subsec:model-editing}
Model editing refers to the paradigm of adding new functionality and behaviors to pre-trained neural models by manipulating the parameters or outputs, without the need of expensive retraining. In particular, \cite{ilharco2023editing} defines editing neural networks based on task vectors, which encode the information necessary to do well on a given task. They obtain such vectors by taking the weights of a model fine-tuned on a task and subtracting the corresponding pre-trained weights. They showcase that performing simple arithmetic operations on these task vectors can adapt a model to a new task or negate an undesirable behavior. In this work, we propose a novel Sequential Model Editing approach that leverages task vectors to expand the abilities of our ASR model without the need to access old data or to introduce any training loss functions and/or additional model parameters. 

At any given stage $t$ of the update of the ASR model, we have access to the model parameters $\theta_{t-1}$ and the new domain/accent with data source $\mathcal{D}_t$. The objective is to learn new set of model parameters $\theta_{t}$ that performs well on the new domain/accent while maintaining its original capabilities. Therefore, the problem can be simplified to learning a task vector $\tau_t$ for the new domain/accent and then leveraging model editing to update the ASR model. The task vector $\tau_t$ will represent information specific to new domain that was missing in previous model checkpoint $\theta_{t-1}$. First, we fine-tune the model on the new data source to arrive at an intermediate model checkpoint $\hat{\theta_t}$. Next, we explore two different ways of creating task vectors from this checkpoint. These two different versions of Sequential Model Editing are defined as follows:

\noindent (1) \textbf{Task Arithmetic \cite{ilharco2023editing}}: In this version, the task vector is defined by simply taking the element-wise difference between $\hat{\theta_t}$ and $\theta_{t-1}$, i.e., $\tau_t = \hat{\theta_t} - \theta_{t-1}$. 

\noindent (2) \textbf{TIES-Merging \cite{yadav2023tiesmerging}}: Since the number of parameters in a model can be substantially large, the dimension of the task vector in the previous version will be equivalently large. Many of the values in this vector will be of low magnitude. 
Therefore, redundant parameters from $\tau_t$ are removed in this version. Specifically, the top-k\% values are retained based on their magnitude, while the bottom (100-k)\% are set to 0. Although the TIES-Merging procedure \cite{yadav2023tiesmerging} involves more complex operations to create the final aggregate task vector when multiple tasks are involved, we omit those details in this work as our sequential model editing procedure involves only a single task at a time.

Finally, the task vector created by either of the two versions is added back to the model to create the final checkpoint $\theta_{t}$:
$$\theta_{t} = \theta_{t-1} + \lambda\cdot(\tau_{t})$$
$\lambda = 1$ corresponds to fine-tuning, which leads to Catastrophic Forgetting, while $\lambda = 0$ means no model update. The optimal $\lambda$ is chosen via held-out validation, balancing these two extremes.

\textbf{Similarities with Model Averaging for ASR:} Weight averaging is a well-known technique in lifelong learning for ASR \cite{vander2023rehearsal, eeckt2022weight}, where a weighted average of a previous and adapted model is computed. Task Arithmetic can be seen as a form of weight averaging, but our model editing approach extends this by incorporating additional operations, similar to TIES-Merging, and other advanced editing techniques (\cite{yu2024language, muqeeth2023soft}).



\section{Experiments}
\subsection{Data}
We use the CommonVoice English ASR data \cite{ardila2019common} partitioned by accents. We adopt the data settings outlined in \cite{eeckt2023rehearsal} and retrieve the data from their open-source GitHub repository\footnote{https://github.com/StevenVdEeckt/online-cl-for-asr}. Our lifelong learning experiments incrementally improve ASR performance across six accents: US, ENG, AUS, IND, SCO, and IRE. The initial model $\theta_0$ is trained on US data ($\mathcal{D}_0$), and accents are added in the order US$\rightarrow$ENG$\rightarrow$AUS$\rightarrow$IND$\rightarrow$SCO$\rightarrow$IRE. This order starts with the largest dataset (US) to create a strong base model, and subsequent accents are sequenced randomly. Dataset specifics are shown in Table \ref{tab:data-stats}.

\begin{table}[]
    \centering
    \begin{tabular}{lll|lll}
      \textbf{Notation} & \textbf{Accent} & \textbf{Country} & \textbf{Train} & \textbf{Dev} & \textbf{Test}\\
       \toprule
        $\mathcal{D}_0$ & US & United States & 470 & 1.4 & 1.6 \\
        $\mathcal{D}_1$ & ENG & England & 152 & 1.2 & 1.2\\
        $\mathcal{D}_2$ & AUS & Australia & 78 & 1 & 1.4\\
        $\mathcal{D}_3$ & IND & India & 104 & 1.3 & 1.6\\
        $\mathcal{D}_4$ & SCO & Scotland & 17 & 1 & 1.3\\
        $\mathcal{D}_5$ & IRE & Ireland & 10 & 1 & 1.4\\
       \bottomrule
    \end{tabular}
    \caption{CV English dataset duration (hrs) per accent.}
    \vspace{-0.5cm}
    \label{tab:data-stats}
\end{table}
\subsection{Model Architecture}
We use a 12-layer CTC Conformer model, incorporating 8 self-attention heads, a 1024-dimensional feedforward layer, and an input/output size of 80, following the approach \cite{gulati2020conformer}. The models are designed to directly predict subword targets, derived from a sentence-piece model trained on initial US dialect $\mathcal{D}_0$, with a total vocabulary size of 512. The initial training on $D_0$ data source spans 60 epochs with a learning rate of 5e-3. Subsequently, for every addition of new data for an accent, the models undergo an additional 10 epochs of training with a reduced learning rate of 5e-4. To enhance ASR inference, a 4-gram language model is trained on combined data from all accents, and it is employed during beam search. We use the ESPnet library \cite{watanabe2018espnet} for ASR and KenLM \cite{heafield-2011-kenlm} for LM training. All models are updated with the Adam optimizer with a weight decay of 0.1.

\textbf{Model Editing:} We assign $\lambda=0.4$ for Task Arithmetic, and for TIES-Merging, we set $\lambda=0.6$ and $k=0.5$ consistently across all time steps. These specific values are determined through evaluation on the development set at stage $t=1$, and we maintain them unchanged for future stages. Although we did not explore varying $\lambda$ or $k$ for each stage here, such an exploration remains a potential avenue for future research.

\begin{table*}[]
    \centering
    \begin{tabular}{l|l|cccccc|ccc}
        \multicolumn{2}{c|}{\textbf{Method}} & \textbf{US} & \textbf{ENG} & \textbf{AUS} & \textbf{IND} & \textbf{SCO} & \textbf{IRE} & \textbf{AWER} & \textbf{WERR (\%)} \\
        \toprule
        \multirow{3}{*}{Baselines} & Fine-tune & 13.2 & 11.5 & 8.9 & 16.9 & 9 & 7.8 & 11.2 & - \\
        & UOE & 12.3 & 10.9 & 8.4 & 15.4 & \textbf{8.2} & 7.5 & 10.5 & 6.3 \\
        & CLRL-Tuning & 12.9 & 12 & 9.7 & 18 & 9.2 & 8 & 11.6 & -3.5 \\
        \hline
        \multirow{2}{*}{Model Editing} & Task Arithmetic & 12.1 & 9.8 & 9 & 14.8 & 9.1 & 6.4 & 10.2 & 9.1 \\
        & TIES-Merging & \textbf{11.3} & \textbf{8.8} & \textbf{8.2} & \textbf{14.2} & 8.8 & \textbf{5.9} & \textbf{9.5} & \textbf{15} \\
        \hline \hline
        \multirow{2}{*}{Oracle} & Sep. Model & 12.9 & 9.8 & 6.3 & 12.1 & 7.7 & 7.1 & 9.3 & 16.8 \\
        & Multi-task & 13 & 9.6 & 6.2 & 13.4 & 7.3 & 7.2 & 9.4 & 15.7 \\
        \bottomrule
    \end{tabular}
    \caption{WER ($\downarrow$) on the CV English testset after learning the six tasks (i.e. accents) in sequence.}
    \vspace{-0.5cm}
    \label{tab:results}
\end{table*}

\subsection{Baselines}
\begin{itemize}
    \item \textit{Fine-tuning}: This involves fine-tuning previous checkpoint on new accent data, and is expected to be highly susceptible to catastrophic forgetting (CF).
    \item \textit{Randomly Layer-wise (CLRL) Tuning \cite{wangclrl}}: This approach suggests randomly fine-tuning only $M < N$ out of $N$ encoder layers on new data while keeping the remaining $N-M$ layers frozen to mitigate CF. We set $N=1$ as it yields optimal results based on the referenced paper.
    \item \textit{Update Only Encoders (UOE) \cite{takashima2022updating}}: This method involves updating only linear layers of Conformer encoders to prevent CF during incremental domain adaptation. Here, linear layers refer specifically to the weight matrices of the Feedforward Network and attention module within a Conformer block.
\end{itemize}

Other conventional LLL methods like Experience Replay \cite{isele2018selective} require access to the old data at every stage and therefore are not directly comparable to our methods (as our methods explicitly aim to relax this requirement). However, we do benchmark even stronger upper bounds (oracle): (1)\textit{Multi-Task} model, which trains the ASR model on the pooled data from all accents $\mathcal{D}_{combined} = \cup_{t=0}^T \mathcal{D}_t$. This helps to better understand the gap between our methods and the best method when all the datasets are available at every stage. (2) \textit{Sep. Model}, which trains separate ASR models for each accent dataset independently and hence uses more parameters than our methods. 

\subsection{Metrics}
We report the WER per task, average WER across seen accents (AWER), and WER reduction (WERR \%) across seen accents compared to the fine-tuning baseline. In this context, "seen" accents refer to those accents for which the corresponding data source has been used in any stage of training. For instance, at time step $t=2$, the AWER is computed as the average of the baseline (US) and the next two accents (ENG, AUS) WER on the test sets $\mathcal{D}_0, \mathcal{D}_1, \mathcal{D}_2$, respectively.

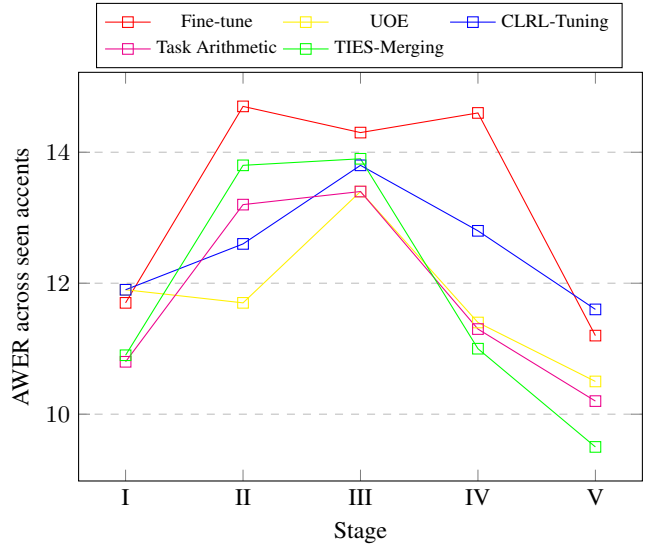
\begin{figure}
\begin{tikzpicture}
\begin{axis}[
    xlabel={Stage},
    ylabel={AWER across seen accents},
    xticklabels={I,II,III,IV,V},
    xtick={1,2,3,4,5},
    width=9cm,height=7cm,
    legend columns=3,
    legend style={font=\scriptsize, at={(.965,1.17)}},
    ymajorgrids=true,
    grid style=dashed,
]

\addplot[
    color=red,
    mark=square,
    ]
    coordinates {
    (1,11.7)(2,14.7)(3,14.3)(4,14.6)(5,11.2)
    };
    \addlegendentry{Fine-tune}

\addplot[
    color=yellow,
    mark=square,
    ]
    coordinates {
    (1,11.9)(2,11.7)(3,13.4)(4,11.4)(5,10.5)
    };
    \addlegendentry{UOE}

\addplot[
    color=blue,
    mark=square,
    ]
    coordinates {
    (1,11.9)(2,12.6)(3,13.8)(4,12.8)(5,11.6)
    };
    \addlegendentry{CLRL-Tuning}

\addplot[
    color=magenta,
    mark=square,
    ]
    coordinates {
    (1,10.8)(2,13.2)(3,13.4)(4,11.3)(5,10.2)
    };
    \addlegendentry{Task Arithmetic}

\addplot[
    color=green,
    mark=square,
    ]
    coordinates {
    (1,10.9)(2,13.8)(3,13.9)(4,11)(5,9.5)
    };
    \addlegendentry{TIES-Merging}
    
\end{axis}
\end{tikzpicture}
\caption{Evolution of WER on seen accents for various approaches as new accents are added incrementally.}
\label{fig:wer-evolution}
\end{figure}

\section{Results}

\subsection{Catastrophic Forgetting - Model Editing vs Baselines}

Table \ref{tab:results} presents the results of sequential lifelong learning experiments post the last step ($t=5$), encompassing exposure to all six English tasks (accents).

The average WER (AWER) across all dialects of the oracle multi-task model achieved through training the ASR model on all data is 9.4, while sequential conventional fine-tuning results in a WER of 11.2. This highlights and underscores the existence of the catastrophic forgetting problem. The recently introduced Update Only Encoders (UOE) method \cite{takashima2022updating} exhibits a 6.3\% WERR improvement over the fine-tuning baseline. However, the CLRL-Tuning method \cite{wangclrl} tends to perform below this baseline for most accents. Both these approaches however demonstrate efficacy in scenarios with fewer tasks (refer to section \ref{subsec:stage-wise-eval}) but experiences degradation with the addition of more tasks.

Notably, both of our sequential editing methods show improvement over the baseline. Task Arithmetic and TIES-Merging yield 9.1\% and 15.0\% WERR, respectively, almost reaching the performance of the oracle upper bound methods. The superiority of TIES-Merging, with lower overall WER than Task Arithmetic, underscores the importance of employing additional steps during merging of task vectors as proposed in \cite{yadav2023tiesmerging}.

\subsection{Stage-wise Analysis}\label{subsec:stage-wise-eval}
We evaluate the performance of various approaches at each time step of task addition, ranging from $t=1$ (introducing $\mathcal{D}_1$) to $t=5$ (introducing $\mathcal{D}_5$). The AWER at each time step is computed, and Figure \ref{fig:wer-evolution} visually represents the results.

Our observations indicate that previously proposed methods, such as UOE and CLRL-Tuning, exhibit superiority over the baseline fine-tuning and even match the performance of our model editing approach in the initial stages up to $t=3$. However, with the incorporation of additional accented data, these approaches start encountering the issue of forgetting, and the model editing approaches consistently outperform all baselines. This underscores the scalability of these model-editing approaches for sequential lifelong learning, with the potential for further enhancements as more accents are introduced.

\subsection{Incremental improvements in Model Editing}
To assess the incremental enhancements in model editing at each time step $t$, we compare the intermediate fine-tuned checkpoint $\hat{\theta}_t$ with the edited checkpoint $\theta_t$, where $\theta_t = \theta_{t-1} + \lambda(\tau_t)$. Table \ref{tab:editing-gains} presents the AWER for various time steps, comparing both the fine-tuned and edited (TIES-Merging) checkpoints. Note that this fine-tuned checkpoint is different than conventional fine-tuning baseline, since the former is fine-tuned on previously edited checkpoint. The table shows that incorporating task vectors consistently enhances performance at all stages compared to fine-tuning, with WERR gains between 1.5-14\%.
\begin{table}[]
    \centering
    \begin{tabular}{c|ccccc}
        \textbf{Model/$T$} & \textbf{$t=1$} & \textbf{$t=2$} & \textbf{$t=3$} & \textbf{$t=4$} & \textbf{$t=5$}\\
        \toprule
        Intermediate ($\hat{\theta_t}$) & 11.7 & 14.2 & 14.1 & 12.8 & 10.8 \\
        Edited ($\theta_t$) & 10.9 & 13.8 & 13.9 & 11 & 9.5 \\
        \hline
        WERR (\%)  & 3.4 & 2.8 & 1.5 & 14 & 12 \\
        \bottomrule
    \end{tabular}
    \caption{AWER at every timestep for intermeditate fine-tuned vs edited checkpoint.}
    \label{tab:editing-gains}
    \vspace{-0.8cm}
\end{table}

\subsection{Choosing the optimal $\lambda$ to mitigate CF}\label{subsec:optimal_lambda}
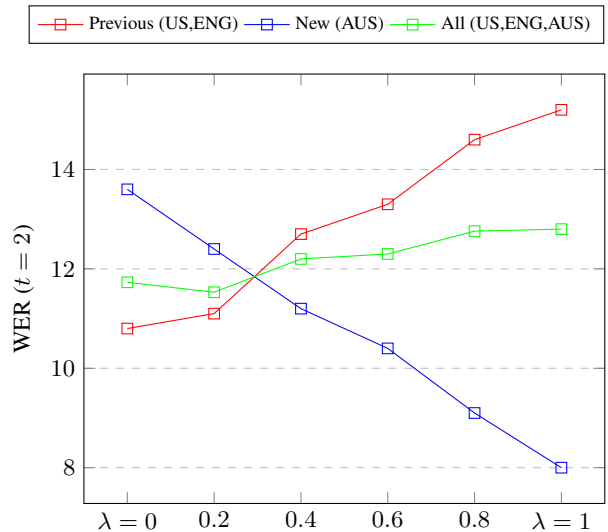
\begin{figure}
\begin{tikzpicture}
\begin{axis}[
    ylabel={WER ($t=2$)},
    xticklabels={$\lambda=0$,$0.2$,$0.4$,$0.6$,$0.8$,$\lambda=1$},
    xtick={0,1,2,3,4,5},
    legend columns=3,
    legend style={font=\scriptsize, at={(0.999,1.16)}},
    ymajorgrids=true,
    grid style=dashed,
]

\addplot[
    color=red,
    mark=square,
    ]
    coordinates {
    (0,10.8)(1,11.1)(2,12.7)(3,13.3)(4,14.6)(5,15.2)
    };
    \addlegendentry{Previous (US,ENG)}

\addplot[
    color=blue,
    mark=square,
    ]
    coordinates {
    (0,13.6)(1,12.4)(2,11.2)(3,10.4)(4,9.1)(5,8)
    };
    \addlegendentry{New (AUS)}

\addplot[
    color=green,
    mark=square,
    ]
    coordinates {
    (0,11.73)(1,11.53)(2,12.2)(3,12.3)(4,12.76)(5,12.8)
    };
    \addlegendentry{All (US,ENG,AUS)}
    
\end{axis}
\end{tikzpicture}
\caption{Variation in WER for previous seen accents vs new accent for $t=2$ for different $\lambda$.}
\label{fig:lambda-variation}
\end{figure}

To analyze the impact of the scaling factor $\lambda$ in the Task Arithmetic technique, we conducted an ablation study by varying $\lambda$ during $t=2$. The results, illustrated in Figure \ref{fig:lambda-variation}, reveal impacts across previous accents, new accents, and the average of both.

Notably, $\lambda=1$, or full fine-tuning, results in catastrophic forgetting for previous accents, leading to the worst performance. Conversely, this setting yields the best performance for the new accent. Intriguingly, a $\lambda$ value of 0.2 emerges as the most optimal, in contrast to the fixed value of 0.4 used for all Task Arithmetic-based model merging. This finding suggests the potential for fine-tuning $\lambda$ differently for each time step and even tailoring it for different task vectors. We leave this avenue for further exploration in future work.

\section{Conclusion}
We address the challenge of adapting Automatic Speech Recognition (ASR) to new domains by introducing Lifelong Learning (LLL) algorithms. Traditional methods face computational inefficiencies and concerns about Catastrophic Forgetting (CF) during fine-tuning. While previous LLL techniques exist, our study propose Sequential Model Editing, a novel approach that does not require previous datasets or additional parameters. Empirical results showcase up to a 15\% Word Error Rate Reduction (WERR) over the fine-tuning baseline and superior efficiency compared to other LLL techniques on the CV English multi-accent dataset. This approach effectively mitigates CF and maintains high performance across diverse domains.

One avenue for future research involves experimenting with varying values of the scaling factor $\lambda$ at different time steps, potentially yielding enhanced improvements, as illustrated in Section \ref{subsec:optimal_lambda}. Another avenue is exploring into the theoretical foundations that contribute to the superior performance of model editing in the context of lifelong learning. Additionally, we intend to explore recently proposed editing techniques, including Drop And REscale \cite{yu2024language} and Soft Merging of Experts \cite{muqeeth2023soft}.
\clearpage

\section{Acknowledgements}
We thank Veera Raghavendra Elluru for his constant feedback during the course of work, as well as rebuttal phase of the paper.
\bibliographystyle{IEEEtran}
\bibliography{mybib}

\end{document}